\documentclass{article}

\usepackage[utf8]{inputenc}
\usepackage[USenglish]{babel}
\usepackage{csquotes}

\usepackage{amsmath,amssymb}
\usepackage{fullpage}
\usepackage{subcaption}
\usepackage{graphicx}

\usepackage{todonotes}

\usepackage{hyperref} 

\usepackage[style=authoryear,maxbibnames=12,maxcitenames=2]{biblatex}
\usepackage[capitalize]{cleveref}

\DeclareMathOperator{\E}{\mathbb{E}}
\DeclareMathOperator{\Var}{Var}
\DeclareMathOperator{\Cov}{Cov}
\DeclareMathOperator{\mmd}{MMD}
\DeclareMathOperator{\mmdsqu}{\widehat{MMD}_U^2}
\newcommand{\tp}{^\mathsf{T}}

\newcommand{\Kxy}{\mathbf{K_{XY}}}
\newcommand{\Kxx}{\mathbf{K_{XX}}}
\newcommand{\Kyy}{\mathbf{K_{YY}}}
\newcommand{\Kxz}{\mathbf{K_{XZ}}}
\newcommand{\Kzz}{\mathbf{K_{ZZ}}}
\newcommand{\Ktxx}{\mathbf{\tilde{K}_{XX}}}
\newcommand{\Ktyy}{\mathbf{\tilde{K}_{YY}}}
\newcommand{\Ktzz}{\mathbf{\tilde{K}_{ZZ}}}

\newcommand{\one}{\mathbf{1}}

\newcommand{\X}{\mathcal{X}}
\newcommand{\R}{\mathbb{R}}
\newcommand{\h}{\mathcal{H}}

\newcommand{\PX}{\mathbb{P}_X}
\newcommand{\PY}{\mathbb{P}_Y}
\newcommand{\PZ}{\mathbb{P}_Z}

\newcommand{\muX}{\mu_X}
\newcommand{\muY}{\mu_Y}
\newcommand{\muZ}{\mu_Z}

\newcommand{\setX}{\mathbf{X}}
\newcommand{\setY}{\mathbf{Y}}
\newcommand{\setZ}{\mathbf{Z}}
\newcommand{\setU}{\mathbf{U}}
\newcommand{\setW}{\mathbf{W}}

\makeatletter

\DeclareRobustCommand{\abs}{\@ifstar\@@abs\@abs}
\newcommand{\@abs}[1]{\lVert #1 \rvert}
\newcommand{\@@abs}[1]{\lVert #1 \rvert}

\DeclareRobustCommand{\norm}{\@ifstar\@@norm\@norm}
\newcommand{\@norm}[1]{\lVert #1 \rVert}
\newcommand{\@@norm}[1]{\lVert #1 \rVert}

\DeclareRobustCommand{\inner}{\@ifstar\@@inner\@inner}
\newcommand{\@inner}[2]{\langle #1, #2 \rangle}
\newcommand{\@@inner}[2]{\langle #1, #2 \rangle}

\makeatother

\title{Unbiased estimators for the variance of MMD estimators}
\author{Danica J.\ Sutherland $\qquad$ Namrata Deka}
\date{November 2022\footnote{Updates since the version of June 2019: fixed a mistake in the coefficient of a leading term (!); allowed for $m$ and $n$ to be distinct; slightly nicer notation with covariance operators; added discussion of the biased estimator.}}

\begin{document}
\maketitle

\begin{abstract}
The maximum mean discrepancy (MMD) is a kernel-based distance between probability distributions useful in many applications \parencite{mmd-jmlr},
bearing a simple estimator with pleasing computational and statistical properties.
Being able to efficiently estimate the variance of this estimator is very helpful to various problems in two-sample testing.
Towards this end,
\textcite{three-sample} used the theory of U-statistics
to derive estimators for the variance of an MMD estimator,
and differences between two such estimators.
Their estimator, however, drops lower-order terms, and is unnecessarily biased.
We show in this note~--~extending and correcting work of \textcite{opt-mmd}~--~%
that we can find a truly unbiased estimator for the actual variance of both the squared MMD estimator and the difference of two correlated squared MMD estimators,
at essentially no additional computational cost.
\end{abstract}

We give only minimal background in this note;
see \textcite{three-sample,opt-mmd} for uses of these estimators.
Since the initial version of this note,
\textcite{liu:deep-testing} has proposed a much simpler biased estimator which sometimes seems to work better,
as in experiments done by \textcite{deka:mmd-b-fair} and theoretical results in their Appendix A; we discuss this in more detail later.

Given a positive semidefinite kernel $k : \X \times \X \to \R$ corresponding to an RKHS $\h$,
there exists a feature map $\varphi : \X \to \h$
such that $k(X, Y) = \inner{\varphi(X)}{\varphi(Y)}$.

The mean embedding \parencite[see e.g.]{mean-embeddings} of a distribution $\PX$ is
$\muX := \E_{X \sim \PX}[ \varphi(X) ] \in \h$,
which exists as long as $\E_{X \sim \PX} \sqrt{k(x, x)} < \infty$;
we assume this is the case for all distributions in this note,
which further allows us to generally exchange expectations with inner products in $\h$ (Bochner integrability).
This technical condition holds automatically for continuous bounded kernels
or for continuous kernels on compact domains,
but must be verified in other situations.

The MMD is the distance between mean embeddings:
\begin{align*}
  \mmd^2(\PX, \PY)
  &= \norm{\muX - \muY}^2
\\&=  \E_{X, X' \sim \PX}\left[ k(X, X') \right]
    + \E_{Y, Y' \sim \PY}\left[ k(Y, Y') \right]
    - 2 \E_{X \sim \PX, Y \sim \PY}\left[ k(X, Y) \right]
.\end{align*}

Suppose we have independent samples
$\setX := \{ X_i \}_{i=1}^m \sim \PX^m$,
$\setY := \{ Y_i \}_{i=1}^m \sim \PY^m$,
$\setZ := \{ Z_i \}_{i=1}^m \sim \PZ^m$.
The following is an unbiased estimator of $\mmd(\PX, \PY)$ with \emph{nearly} minimal variance among unbiased estimators \parencite{mmd-jmlr}:
\begin{equation} \label{eq:mmd-est}
    \mmdsqu(\setX, \setY)
    := \frac{1}{m (m - 1)} \sum_{i \ne j}^m \left[
        k(X_i, X_j) + k(Y_i, Y_j) - k(X_i, Y_j) - k(X_j, Y_i)
    \right]
.\end{equation}
Compared to the MVUE, terms of the form $k(X_i, Y_i)$ are dropped.
This estimator, however, is a $U$ statistic,
for which there is well-established theory \parencite[Chapter~5]{serfling},
including expressions for the variance.

In this note, we first employ that theory to derive expressions for the variance in terms of various expectations of inner products in $\h$ (\cref{sec:var-expressions}).
Then, in \cref{sec:estimators}, we derive unbiased estimators for these expressions,
yielding the final results \eqref{eq:final-mmd-est} and \eqref{eq:final-diff-est}
which are unbiased variance estimators which can be evaluated in the same $\mathcal{O}(m^2)$ time
it takes to evaluate \eqref{eq:mmd-est}.

Note that, in applications such as used by \textcite{opt-mmd,liu:deep-testing},
we wish to estimate not just $\Var[\mmdsqu]$ but actually $\frac{\mmd}{\sqrt{\Var[\mmdsqu]}}$.
An unbiased estimator for the variance does \emph{not} result in an unbiased estimator for $\frac{\mmd}{\sqrt{\Var[\mmdsqu]}}$;
in fact, \textcite{deka:mmd-b-fair} prove that estimator which is unbiased for that quantity exists.
For most use cases, then, the biased estimator of \cref{sec:biased} is much simpler
and can in fact perform better,
as speculated by \textcite{liu:deep-testing}
and confirmed in inital experiments by \textcite{deka:mmd-b-fair}.

\section{Variance expressions} \label{sec:var-expressions}

We will first derive expressions for the variances of
$\mmdsqu(\setX, \setY)$
and $\mmdsqu(\setX, \setY) - \mmdsqu(\setX, \setZ)$.
This section is quite similar to Appendix A of \textcite{three-sample},
but avoids unnecessarily dropping lower-order terms
(which provides almost no computational advantage, and may harm the accuracy for small sample sizes, although it does make for a less tedious derivation).
The result \eqref{eq:final-mmd-exp} of \cref{sec:mmd-expression} is identical to that of \textcite{opt-mmd}.

A useful tool in stating these results is the \emph{uncentred covariance operator}
\[
    C_X = \E_{X \sim \PX}\big[ \varphi(X) \otimes \varphi(X) \big]
,\]
where $a \otimes b$ is the outer product,
a linear operator given by
$[a \otimes b] x = a \inner{b}{x}$.

\subsection{Variance of the MMD estimator} \label{sec:mmd-expression}

Let $U_i$ denote the pair $(X_i, Y_i)$,
and define the function
\[
    h(U_1, U_2) := k(X_1, X_2) + k(Y_1, Y_2) - k(X_1, Y_2) - k(X_2, Y_1)
.\]
Then \[
\mmdsqu(\setX, \setY) = \frac{1}{m (m-1)} \sum_{i \ne j} h(U_i, U_j)
,\]
which is a $U$-statistic on the joint data $\setU$.
Thus standard results \parencite[Section 5.2.1, Lemma A]{serfling}
give us
\begin{equation}
    \Var\left[ \mmdsqu(\setX, \setY) \right]
    = V_m
    := \frac{4 (m-2)}{m (m-1)} \zeta_1 + \frac{2}{m (m-1)} \zeta_2
\label{eq:mmd:var-zetas}
,\end{equation} where
\[
    \zeta_1 := \Var_{U_1}\left[ \E_{U_2}\left[ h(U_1, U_2) \right] \right],
    \qquad
    \zeta_2 := \Var_{U_1, U_2}\left[ h(U_1, U_2) \right]
.\]

The first-order term $\zeta_1$ is:
\begin{align*}
    \zeta_1
  &=
    \Var_{U_1}\left[ \E_{U_2}\left[ h(U_1, U_2) \right] \right]
\\&=
    \Var_{X_1, Y_1}\left[
      \E_{X_2} k(X_1, X_2)
    + \E_{Y_2} k(Y_1, Y_2)
    - \E_{Y_2} k(X_1, Y_2)
    - \E_{X_2} k(X_2, Y_1)
    \right]
\\&=
    \Var_{X, Y}\left[
      \inner{\varphi(X)}{\muX}
    + \inner{\varphi(Y)}{\muY}
    - \inner{\varphi(X)}{\muY}
    - \inner{\muX}{\varphi(Y)}
    \right]
\\&=
    \Var\left[ \inner{\varphi(X)}{\muX} \rangle \right]
  + \Var\left[ \inner{\varphi(Y)}{\muY} \rangle \right]
  + \Var\left[ \inner{\varphi(X)}{\muY} \rangle \right]
  + \Var\left[ \inner{\muX}{\varphi(Y)} \rangle \right]
\\&\quad
  - 2 \Cov\left( \inner{\varphi(X)}{\muX}, \inner{\varphi(X)}{\muY} \right)
  - 2 \Cov\left( \inner{\varphi(Y)}{\muY}, \inner{\muX}{\varphi(Y)} \right)
.\end{align*}
Noting that
\begin{align*}
    \Var\left[ \inner{\varphi(A)}{\mu_B} \right]
  &=
    \E\left[ \inner{\varphi(A)}{\mu_B}^2 \right]
  - \inner{\mu_A}{\mu_B}^2
\\&=
    \inner{\mu_B}{C_A \mu_B}
    - \inner{\mu_A}{\mu_B}^2
\\
    \Cov\left( \inner{\varphi(A)}{\mu_B}, \inner{\varphi(A)}{\mu_C} \right)
  &=
    \E\left[ \inner{\varphi(A)}{\mu_B} \inner{\varphi(A)}{\mu_C} \right]
  - \inner{\mu_A}{\mu_B} \inner{\mu_A}{\mu_C}
\\&=
    \inner{\mu_B}{C_A \mu_C}
    - \inner{\mu_A}{\mu_B}
      \inner{\mu_A}{\mu_C}
\end{align*}
we have
\begin{align*}
    \zeta_1
&=
    \inner{\muX}{C_X \muX} - \inner{\muX}{\muX}^2
\\&\quad
  + \inner{\muY}{C_Y \muY} - \inner{\muY}{\muY}^2
\\&\quad
  + \inner{\muY}{C_X \mu_Y} - \inner{\muX}{\muY}^2
\\&\quad
  + \inner{\muX}{C_Y \muX} - \inner{\muY}{\muX}^2
\\&\quad
  - 2 \inner{\muX}{C_X \muY}
  + 2 \inner{\muX}{\muX} \inner{\muX}{\muY}
\\&\quad
 - 2 \inner{\mu_Y}{C_Y \mu_X}
 + 2 \inner{\muY}{\muY} \inner{\muX}{\muY}
.\end{align*}

We can similarly compute the second-order term $\zeta_2$ as:
\begin{align*}
\zeta_2
  &= \Var\left[ h(U_1, U_2) \right]
\\&= \Var\left[ k(X_1, X_2) + k(Y_1, Y_2) - k(X_1, Y_2) - k(X_2, Y_1) \right]
\\&=
    \Var\left[ k(X_1, X_2) \right]
  + \Var\left[ k(Y_1, Y_2) \right]
  + \Var\left[ k(X_1, Y_2) \right]
  + \Var\left[ k(X_2, Y_1) \right]
\\&\quad
  - 2 \Cov\left( k(X_1, X_2), k(X_1, Y_2) \right)
  - 2 \Cov\left( k(X_1, X_2), k(X_2, Y_1) \right)
\\&\quad
  - 2 \Cov\left( k(Y_1, Y_2), k(X_1, Y_2) \right)
  - 2 \Cov\left( k(Y_1, Y_2), k(X_2, Y_1) \right)
\\&=
    \Var\left[ k(X_1, X_2) \right]
  + \Var\left[ k(Y_1, Y_2) \right]
  + 2 \Var\left[ k(X, Y) \right]
\\&\quad
  - 4 \Cov\left( k(X_1, X_2), k(X_1, Y) \right)
  - 4 \Cov\left( k(Y_1, Y_2), k(Y_1, X) \right)
\\&=
    \E\left[ k(X_1, X_2)^2 \right] - \inner{\muX}{\muX}^2
  + \E\left[ k(Y_1, Y_2)^2 \right] - \inner{\muY}{\muY}^2
  + 2 \E\left[ k(X, Y)^2 \right] - 2 \inner{\muX}{\muY}^2
\\&\quad
  - 4 \inner{\muX}{C_X \muY}
  + 4 \inner{\muX}{\muX} \inner{\muX}{\muY}
  - 4 \inner{\muY}{C_Y \muX}
  + 4 \inner{\muY}{\muY} \inner{\muX}{\muY}
.\end{align*}

Combining the two yields
an expression for $V_m = \Var[\mmdsqu(\setX, \setY)]$
of
\begin{align*}
    V_m
  &= \frac{2}{m (m-1)} \Big[ 2 (m-2) \zeta_1 + \zeta_2 \Big]
\\&= \frac{2}{m (m-1)} \Bigg[
    2 (m-2) \inner{\muX}{C_X \muX}
  - 2 (m-2) \inner{\muX}{\muX}^2
\\&\qquad\qquad\qquad
  + 2 (m-2) \inner{\muY}{C_Y \muY}
  - 2 (m-2) \inner{\muY}{\muY}^2
\\&\qquad\qquad\qquad
  + 2 (m-2) \inner{\muY}{C_X \muY}
  - 2 (m-2) \inner{\muX}{\muY}^2
\\&\qquad\qquad\qquad
  + 2 (m-2) \inner{\muX}{C_Y \muX}
  - 2 (m-2) \inner{\muY}{\muX}^2
\\&\qquad\qquad\qquad
  - 4 (m-2) \inner{\muX}{C_X \muY}
  + 4 (m-2) \inner{\muX}{\muX} \inner{\muX}{\muY}
\\&\qquad\qquad\qquad
  - 4 (m-2) \inner{\muY}{C_Y \muX}
  + 4 (m-2) \inner{\muY}{\muY} \inner{\muX}{\muY}
\\&\qquad\qquad\qquad
  + \E\left[ k(X_1, X_2)^2 \right] - \inner{\muX}{\muX}^2
\\&\qquad\qquad\qquad
  + \E\left[ k(Y_1, Y_2)^2 \right] - \inner{\muY}{\muY}^2
\\&\qquad\qquad\qquad
  + 2 \E\left[ k(X, Y)^2 \right] - 2 \inner{\muX}{\muY}^2
\\&\qquad\qquad\qquad
  - 4 \inner{\muX}{C_X \muY}
  + 4 \inner{\muX}{\muX} \inner{\muX}{\muY}
\\&\qquad\qquad\qquad
  - 4 \inner{\muY}{C_Y \muX}
  + 4 \inner{\muY}{\muY} \inner{\muX}{\muY}
  \Bigg]
\end{align*}
and so, simplifying,
\begin{equation} \label{eq:final-mmd-exp}
\begin{aligned}
    V_m
  &= \frac{2}{m (m-1)} \Bigg[
    2 (m-2) \inner{\muX}{C_X \muX}
  - (2 m - 3) \inner{\muX}{\muX}^2
\\&\qquad\qquad\qquad
  + 2 (m-2) \inner{\muY}{C_Y \muY}
  - (2 m - 3) \inner{\muY}{\muY}^2
\\&\qquad\qquad\qquad
  + 2 (m-2) \big( \inner{\muY}{C_X \muY} + \inner{\muX}{C_Y \muX} \big)
  - (4 m - 6) \inner{\muX}{\muY}^2
\\&\qquad\qquad\qquad
  - 4 (m-1) \inner{\muX}{C_X \muY}
  + 4 (m-1) \inner{\muX}{\muX} \inner{\muX}{\muY}
\\&\qquad\qquad\qquad
  - 4 (m-1) \inner{\muY}{C_Y \muX}
  + 4 (m-1) \inner{\muY}{\muY} \inner{\muX}{\muY}
\\&\qquad\qquad\qquad
  + \E\left[ k(X_1, X_2)^2 \right]
  + \E\left[ k(Y_1, Y_2)^2 \right]
  + 2 \E\left[ k(X, Y)^2 \right]
  \Bigg]
.\end{aligned}
\end{equation}

We can see that, compared to $\zeta_1$,
\eqref{eq:final-mmd-exp} mostly just tweaks constants.
The only new terms are the expectations of squared kernels,
which as we'll see later will in fact not introduce any new computational expense;
their estimators combine with terms needed to correct biases in the other terms.

\subsection{Variance of the difference of MMD estimators} \label{sec:diff-expression}

Let $W_i := (X_i, Y_i, Z_i)$, and
\[
f(W_1, W_2) := 
\left( k(Y_1, Y_2) - k(X_1, Y_2) - k(X_2, Y_1) \right)
- \left( k(Z_1, Z_2) - k(X_1, Z_2) - k(X_2, Z_1) \right)
.\]
Then
\[
  \mmdsqu(\setX, \setY) - \mmdsqu(\setX, \setZ)
  = \frac{1}{m (m-1)} \sum_{i \ne j} f(W_i, W_j)
\]
is a $U$-statistic on $\setW$, so again
\begin{gather*}
  \Var[\mmdsqu(\setX, \setY) - \mmdsqu(\setX, \setZ)]
  = \nu_m
  := \frac{4 (m-2)}{m (m-1)} \xi_1 + \frac{2}{m (m-1)} \xi_2
\\
  \xi_1 := \Var_{W_1}\left[ \E_{W_2}\left[ f(W_1, W_2) \right] \right]
  \qquad
  \xi_2 := \Var_{W_1, W_2}\left[ f(W_1, W_2) \right]
.\end{gather*}

We can proceed as before, but with more terms.
The first-order term $\xi_1$ is
\begin{align*}
     \xi_1
  &= \Var_{W_1}\left[ \E_{W_2}\left[ f(W_1, W_2 )\right] \right]
\\&=
  \Var_{X, Y, Z}\left[
    \inner{\varphi(Y)}{\muY}
  - \inner{\varphi(X)}{\muY}
  - \inner{\muX}{\varphi(Y)}
  - \inner{\varphi(Z)}{\muZ}
  + \inner{\varphi(X)}{\muZ}
  + \inner{\muX}{\varphi(Z)}
  \right]
\\&=
    \Var\left[ \inner{\varphi(X)}{\muY} \right]
  + \Var\left[ \inner{\varphi(X)}{\muZ} \right]
  + \Var\left[ \inner{\varphi(Y)}{\muX} \right]
  + \Var\left[ \inner{\varphi(Y)}{\muY} \right]
  + \Var\left[ \inner{\varphi(Z)}{\muX} \right]
  + \Var\left[ \inner{\varphi(Z)}{\muZ} \right]
\\&\qquad
  - 2 \Cov\left( \inner{\varphi(X)}{\muY}, \inner{\varphi(X)}{\muZ} \right)
  - 2 \Cov\left( \inner{\varphi(Y)}{\muX}, \inner{\varphi(Y)}{\muY} \right)
  - 2 \Cov\left( \inner{\varphi(Z)}{\muX}, \inner{\varphi(Z)}{\muZ} \right)
\\&=
    \inner{\muY}{C_X \muY}
  - \inner{\muX}{\muY}^2
\\&\qquad
  + \inner{\muZ}{C_X \muZ}
  - \inner{\muX}{\muZ}^2
\\&\qquad
  + \inner{\muX}{C_Y \muX}
  - \inner{\muY}{\muX}^2
\\&\qquad
  + \inner{\muY}{C_Y \muY}
  - \inner{\muY}{\muY}^2
\\&\qquad
  + \inner{\muX}{C_Z \muX}
  - \inner{\muZ}{\muX}^2
\\&\qquad
  + \inner{\muZ}{C_Z \muZ}
  - \inner{\muZ}{\muZ}^2
\\&\qquad
  - 2 \inner{\muY}{C_X \muZ}
  + 2 \inner{\muX}{\muY} \inner{\muX}{\muZ}
\\&\qquad
  - 2 \inner{\muX}{C_Y \muY}
  + 2 \inner{\muX}{\muY} \inner{\muY}{\muY}
\\&\qquad
  - 2 \inner{\muX}{C_Z \muZ}
  + 2 \inner{\muX}{\muZ} \inner{\muZ}{\muZ}
.\end{align*}

The second-order term $\xi_2$ is
\begin{align*}
    \xi_2
  &=
    \Var\left[ - k(X_1, Y_2) - k(X_2, Y_1) + k(X_1, Z_2) + k(X_2, Z_1) + k(Y_1, Y_2) - k(Z_1, Z_2) \right]
\\&=
    \Var\left[ k(X_1, Y_2) \right]
  + \Var\left[ k(X_2, Y_1) \right]
  + \Var\left[ k(X_1, Z_2) \right]
  + \Var\left[ k(X_2, Z_1) \right]
  + \Var\left[ k(Y_1, Y_2) \right]
  + \Var\left[ k(Z_1, Z_2) \right]
\\&\qquad
  - 2 \Cov\left( k(X_1, Y_2), k(X_1, Z_2) \right)
  - 2 \Cov\left( k(X_1, Y_2), k(Y_1, Y_2) \right)
\\&\qquad
  - 2 \Cov\left( k(X_2, Y_1), k(X_2, Z_1) \right)
  - 2 \Cov\left( k(X_2, Y_1), k(Y_1, Y_2) \right)
\\&\qquad
  - 2 \Cov\left( k(X_1, Z_2), k(Z_1, Z_2) \right)
  - 2 \Cov\left( k(X_2, Z_1), k(Z_1, Z_2) \right)
\\&=
    2 \Var\left[ k(X, Y) \right]
  + 2 \Var\left[ k(X, Z) \right]
  + \Var\left[ k(Y_1, Y_2) \right]
  + \Var\left[ k(Z_1, Z_2) \right]
\\&\qquad
  - 4 \Cov\left( k(X, Y), k(X, Z) \right)
  - 4 \Cov\left( k(X, Y_1), k(Y_1, Y_2) \right)
  - 4 \Cov\left( k(X, Z_1), k(Z_1, Z_2) \right)
\\&=
    2 \E\left[ k(X, Y)^2 \right]
  - 2 \inner{\muX}{\muY}^2
\\&\qquad
  + 2 \E\left[ k(X, Z)^2 \right]
  - 2 \inner{\muX}{\muZ}^2
\\&\qquad
  + \E\left[ k(Y_1, Y_2)^2 \right]
  - \inner{\muY}{\muY}^2
\\&\qquad
  + \E\left[ k(Z_1, Z_2)^2 \right]
  - \inner{\muZ}{\muZ}^2
\\&\qquad
  - 4 \inner{\muY}{C_X \muZ}
  + 4 \inner{\muX}{\muY} \inner{\muX}{\muZ}
\\&\qquad
  - 4 \inner{\muX}{C_Y \muY}
  + 4 \inner{\muX}{\muY} \inner{\muY}{\muY}
\\&\qquad
  - 4 \inner{\muX}{C_Z \muZ}
  + 4 \inner{\muX}{\muZ} \inner{\muZ}{\muZ}
.\end{align*}

Combining the two
gives an expression for $\Var\left[\mmdsqu(\setX, \setY) - \mmdsqu(\setX, \setZ)\right]$ of
\begin{align*}
    \nu_m
  &= \frac{2}{m (m-1)} \Big[ 2 (m-2) \xi_1 + \xi_2 \Big]
\\&= \frac{2}{m (m-1)} \Bigg[
    2(m-2) \E\left[ \inner{\varphi(X)}{\muY}^2 \right]
  - 2(m-2) \inner{\muX}{\muY}^2
\\&\qquad\qquad\qquad
  + 2(m-2) \E\left[ \inner{\varphi(X)}{\muZ}^2 \right]
  - 2(m-2) \inner{\muX}{\muZ}^2
\\&\qquad\qquad\qquad
  + 2(m-2) \E\left[ \inner{\varphi(Y)}{\muX}^2 \right]
  - 2(m-2) \inner{\muY}{\muX}^2
\\&\qquad\qquad\qquad
  + 2(m-2) \E\left[ \inner{\varphi(Y)}{\muY}^2 \right]
  - 2(m-2) \inner{\muY}{\muY}^2
\\&\qquad\qquad\qquad
  + 2(m-2) \E\left[ \inner{\varphi(Z)}{\muX}^2 \right]
  - 2(m-2) \inner{\muZ}{\muX}^2
\\&\qquad\qquad\qquad
  + 2(m-2) \E\left[ \inner{\varphi(Z)}{\muZ}^2 \right]
  - 2(m-2) \inner{\muZ}{\muZ}^2
\\&\qquad\qquad\qquad
  - 4(m-2) \E\left[ \inner{\varphi(X)}{\muY} \inner{\varphi(X)}{\muZ} \right]
  + 4(m-2) \inner{\muX}{\muY} \inner{\muX}{\muZ}
\\&\qquad\qquad\qquad
  - 4(m-2) \E\left[ \inner{\varphi(Y)}{\muX} \inner{\varphi(Y)}{\muY} \right]
  + 4(m-2) \inner{\muX}{\muY} \inner{\muY}{\muY}
\\&\qquad\qquad\qquad
  - 4(m-2) \E\left[ \inner{\varphi(Z)}{\muX} \inner{\varphi(Z)}{\muZ} \right]
  + 4(m-2) \inner{\muX}{\muZ} \inner{\muZ}{\muZ}
\\&\qquad\qquad\qquad
  + 2 \E\left[ k(X, Y)^2 \right]
  - 2 \inner{\muX}{\muY}^2
\\&\qquad\qquad\qquad
  + 2 \E\left[ k(X, Z)^2 \right]
  - 2 \inner{\muX}{\muZ}^2
\\&\qquad\qquad\qquad
  + \E\left[ k(Y_1, Y_2)^2 \right]
  - \inner{\muY}{\muY}^2
\\&\qquad\qquad\qquad
  + \E\left[ k(Z_1, Z_2)^2 \right]
  - \inner{\muZ}{\muZ}^2
\\&\qquad\qquad\qquad
  - 4 \E\left[ \inner{\varphi(X)}{\muY} \inner{\varphi(X)}{\muZ} \right]
  + 4 \inner{\muX}{\muY} \inner{\muX}{\muZ}
\\&\qquad\qquad\qquad
  - 4 \E\left[ \inner{\varphi(Y)}{\muX} \inner{\varphi(Y)}{\muY} \right]
  + 4 \inner{\muX}{\muY} \inner{\muY}{\muY}
\\&\qquad\qquad\qquad
  - 4 \E\left[ \inner{\varphi(Z)}{\muX} \inner{\varphi(Z)}{\muZ} \right]
  + 4 \inner{\muX}{\muZ} \inner{\muZ}{\muZ}
  \Bigg]
\end{align*}
so that
\begin{equation} \label{eq:final-diff-exp}
\begin{aligned}
    \nu_m
  &= \frac{2}{m (m-1)} \Bigg[
    2(m-2) \Big[
      \inner{\muY}{C_X \muY}
    + \inner{\muZ}{C_X \muZ}
    + \inner{\muX}{C_Y \muX}
    + \inner{\muX}{C_Z \muX}
    \Big]
\\&\qquad\qquad\qquad
  + 2(m-2) \Big[
      \inner{\muY}{C_Y \muY}
    + \inner{\muZ}{C_Z \muZ}
    \Big]
\\&\qquad\qquad\qquad
  - 2(2m-3) \left[
      \inner{\muX}{\muY}^2
    + \inner{\muX}{\muZ}^2
    \right]
  - (2m-3) \left[
      \inner{\muY}{\muY}^2
    + \inner{\muZ}{\muZ}^2
    \right]
\\&\qquad\qquad\qquad
  + 4(m-1) \left[
      \inner{\muX}{\muY} \inner{\muX}{\muZ}
    + \inner{\muX}{\muY} \inner{\muY}{\muY}
    + \inner{\muX}{\muZ} \inner{\muZ}{\muZ}
  \right]
\\&\qquad\qquad\qquad
  - 4(m-1) \Big[
      \inner{\muY}{C_X \muZ}
    + \inner{\muX}{C_Y \muY}
    + \inner{\muX}{C_Z \muZ}
  \Big]
\\&\qquad\qquad\qquad
  + 2 \E\left[ k(X, Y)^2 \right]
  + 2 \E\left[ k(X, Z)^2 \right]
  + \E\left[ k(Y_1, Y_2)^2 \right]
  + \E\left[ k(Z_1, Z_2)^2 \right]
  \Bigg]
.\end{aligned}
\end{equation}

\section{Estimators of terms} \label{sec:estimators}

The expressions above are population quantities that give the exact variance $V_m$ of an estimator based on $m$ samples.
We will now derive estimators of those quantities
based on $n$ samples.
We often might have $m = n$,
but we also might not,
e.g.\ if we wish to get a rough estimate of the power of a test based on $m = 2\,000$ samples using only a minibatch of size $n = 64$,
or if we want a highly accurate understanding of the variance of an estimator on $m = 20$ samples using $n=10\,000$ samples.

\subsection{Biased estimator} \label{sec:biased}
We first note that \textcite{liu:deep-testing} proposed a simple biased estimator for $\zeta_1$
of \eqref{eq:mmd:var-zetas}
based on
\begin{align*}
     \zeta_1
  &= \Var_{U_1}\big[ \E_{U_2}[ h(U_1, U_2) ] \big]
\\&= \E_{U_1}\big[ \E_{U_2}[ h(U_1, U_2) ]^2 \big]
   - \E_{U_1,U_2}\big[ h(U_1, U_2) \big]^2
,\end{align*}
so that $\zeta_1$ may be estimated by
\[
    \zeta_1
    \approx
    \frac{1}{n^3} \sum_{i=1}^n \left( \sum_{j=1}^n h(U_i, U_j) \right)^2
    - \frac{1}{n^4} \left( \sum_{i=1}^n \sum_{j=1}^n h(U_i, U_j) \right)^2
;\]
multiplying by $4 / m$ is then a reasonable estimate of the variance $V_m$.
One could also similarly estimate $\zeta_2$ and plug into \eqref{eq:mmd:var-zetas},
but the $1/n$ bias in this estimator \parencite[Lemma 18]{liu:deep-testing} is of the same order as the contribution of the $\zeta_2$ term when $m = n$.

\subsection{Sub-expressions} \label{sec:sub-estimators}
We will now, instead,
derive an unbiased estimator for $V_m$
by finding an unbiased estimator for each of the various terms in the variance results of \cref{sec:var-expressions}.

Define an $n \times n$ matrix $\Kxy$ by $(\Kxy)_{ij} = k(X_i, Y_j)$,
and $\Kxz$, $\Kxx$, $\Kyy$, $\Kzz$ similarly.
Let $\Ktxx$, $\Ktyy$, $\Ktzz$ be $\Kxx$, $\Kyy$, $\Kzz$ with their diagonals set to zero.
Let $\one$ be the $n$-vector of all ones.
We'll also use the falling factorial notation
$(n)_k := n (n-1) \cdots (n - k + 1)$.

For unbiased estimators, the important thing is to subtract off elements of sums which share data points. For example,
\begin{gather*}
\inner{\muX}{\muY}
= \inner{\E_X \varphi(X)}{\E_Y \varphi(Y)}
= \E_{X, Y} k(X, Y)
\approx \frac{1}{n^2} \sum_{i,j} k(X_i, Y_j)
   = \frac{1}{n^2} \one\tp \Kxy \one
\\
\inner{\muX}{\muX}
= \inner{\E_X \varphi(X)}{\E_{X'} \varphi(X')}
= \E_{X, X'} k(X, X')
\approx \frac{1}{n (n-1)} \sum_{i \ne j} k(X_i, X_j)
   = \frac{1}{n (n-1)} \one\tp \Ktxx \one
.\end{gather*}
It is also important to do so for products of these terms:
this caused the bias present in the publication version of \textcite{opt-mmd}.
For instance,
the square of an unbiased estimator for $\inner{\muX}{\muY}$
is not unbiased for $\inner{\muX}{\muY}^2$,
but the following is:
\begin{align*}
     \inner{\muX}{\muY}^2
  &= \inner{\muX}{\muY} \inner{\muX}{\muY}
   = \E_{X, X', Y, Y'}\left[ k(X, Y) k(X', Y') \right]
\\&\approx
    \frac{1}{n^2} \sum_{i,j} k(X_i, Y_j)
      \frac{1}{(n-1)^2} \sum_{i' \ne i} \sum_{j' \ne j} k(X_{i'}, Y_{j'})
\\&=
    \frac{1}{n^2 (n-1)^2} \sum_{i,j} k(X_i, Y_j) \left[
      \sum_{i',j'} k(X_{i'}, Y_{j'})
    - \sum_{i'} k(X_{i'}, Y_j)
    - \sum_{j'} k(X_i, Y_{j'})
    + k(X_i, Y_j)
    \right]
\\&= \frac{1}{n^2 (n-1)^2} \left[
      \sum_{i,j,i',j'} (\Kxy)_{ij} (\Kxy)_{i'j'}
    - \sum_{i,j,i'} (\Kxy)_{ij} (\Kxy)_{i'j}
    - \sum_{i,j,j'} (\Kxy)_{ij} (\Kxy)_{ij'}
    + \sum_{i,j} (\Kxy)_{ij}^2
    \right]
\\&= \frac{1}{n^2 (n-1)^2} \left[
      \left( \one\tp \Kxy \one \right)^2
    - \one\tp \Kxy \Kxy\tp \one
    - \one\tp \Kxy\tp \Kxy \one
    + \norm{\Kxy}_F^2
    \right]
\\&= \frac{1}{n^2 (n-1)^2} \left[
      \left( \one\tp \Kxy \one \right)^2
    - \norm{\Kxy\tp \one}^2
    - \norm{\Kxy \one}^2
    + \norm{\Kxy}_F^2
    \right]
.\end{align*}
Similarly,
\begin{align*}
     \inner{\muX}{\muX}^2
  &= \E_{X_1, X_2, X_3, X_4}\left[ k(X_1, X_2) k(X_3, X_4) \right]
\\&\approx
    \frac{1}{(n)_4} \sum_{i} \sum_{j \ne i} k(X_i, X_j) \sum_{a \notin \{i, j\}} \sum_{b \notin \{i, j, a\}} k(X_a, X_b)
\\&=
    \frac{1}{(n)_4} \sum_{ij} (\Ktxx)_{ij} \sum_{a, b \notin \{i, j\}} (\Ktxx)_{ab}
\\&=
    \frac{1}{(n)_4} \sum_{ij} (\Ktxx)_{ij} \Bigg[
      \sum_{a b} (\Ktxx)_{ab}
    - \sum_a (\Ktxx)_{ai}
    - \sum_a (\Ktxx)_{aj}
    - \sum_b (\Ktxx)_{ib}
    - \sum_b (\Ktxx)_{jb}
\\&\qquad\qquad\qquad\qquad\qquad
    + (\Ktxx)_{ii}
    + (\Ktxx)_{ij}
    + (\Ktxx)_{ji}
    + (\Ktxx)_{jj}
    \Bigg]
\\&=
    \frac{1}{(n)_4} \sum_{ij} (\Ktxx)_{ij} \Bigg[
      \sum_{a b} (\Ktxx)_{ab}
    - 2 \sum_a (\Ktxx)_{ai}
    - 2 \sum_a (\Ktxx)_{aj}
    + 2 (\Ktxx)_{ij}
    \Bigg]
\\&=
    \frac{1}{(n)_4} \Bigg[
      \sum_{i j a b} (\Ktxx)_{ij} (\Ktxx)_{ab}
    - 2 \sum_{i j a} (\Ktxx)_{ij} (\Ktxx)_{ai}
    - 2 \sum_{i j a} (\Ktxx)_{ij} (\Ktxx)_{aj}
    + 2 \sum_{ij} (\Ktxx)_{ij}^2
    \Bigg]
\\&=
    \frac{1}{(n)_4} \Bigg[
      \left( \one\tp \Ktxx \one \right)^2
    - 4 \norm{\Ktxx \one}^2
    + 2 \norm{\Ktxx}_F^2
    \Bigg]
.\end{align*}
We also need
\begin{align*}
    \inner{\muX}{\muX} \inner{\muX}{\muY}
  &= \E_{X_1, X_2, X_3, Y}\left[ k(X_1, X_2) k(X_3, Y) \right]
\\&\approx
    \frac{1}{n (n)_3} \sum_i \sum_{j \ne i} k(X_i, X_j)
      \sum_{\ell \notin \{i, j\}} \sum_a k(X_\ell, Y_a)
\\&=
    \frac{1}{n (n)_3} \sum_{ij} (\Ktxx)_{ij} \left[
      \sum_{\ell a} (\Kxy)_{\ell a}
    - \sum_a (\Kxy)_{i a}
    - \sum_a (\Kxy)_{j a}
    \right]
\\&=
    \frac{1}{n (n)_3} \left[
      \sum_{i j \ell a} (\Ktxx)_{ij} (\Kxy)_{\ell a}
    - \sum_{i j a} (\Ktxx)_{ij} (\Kxy)_{i a}
    - \sum_{i j a} (\Ktxx)_{ij} (\Kxy)_{j a}
    \right]
\\&=
    \frac{1}{n (n)_3} \left[
      \one\tp \Ktxx \one \one\tp \Kxy \one
    - 2 \, \one\tp \Ktxx \Kxy \one
    \right]
\end{align*}
and
\begin{align*}
    \inner{\muX}{\muY} \inner{\muX}{\muZ}
  &\approx
    \frac{1}{n^3 (n-1)} \sum_i \sum_a \sum_{j \ne i} \sum_b k(X_i, Y_a) k(X_j, Z_b)
\\&= \frac{1}{n^3 (n-1)} \left[
      \sum_{i j a b} (\Kxy)_{ia} (\Kxz)_{jb}
    - \sum_{i a b} (\Kxy)_{ia} (\Kxz)_{ib}
    \right]
\\&= \frac{1}{n^3 (n-1)} \left[
      \one\tp \Kxy \one \one\tp \Kxz \one
    - \one\tp \Kxy\tp \Kxz \one
    \right]
.\end{align*}

We also need some similar terms with $C_X$, which shares a $\varphi(X)$:
\begin{align*}
    \inner{\muX}{C_X \muX}
  &\approx
    \frac{1}{(n)_3} \sum_i \sum_{j \ne i} \sum_{\ell \notin \{i, j\}} k(X_i, X_j) k(X_i, X_\ell)
\\&= \frac{1}{(n)_3} \sum_{ij} \sum_{\ell \notin \{i, j\}} (\Ktxx)_{ij} k(X_i, X_\ell)
\\&= \frac{1}{(n)_3} \sum_{ij} \sum_{\ell \ne j} (\Ktxx)_{ij} (\Ktxx)_{i\ell}
\\&= \frac{1}{(n)_3} \left[
      \sum_{ij\ell} (\Ktxx)_{ij} (\Ktxx)_{i\ell}
    - \sum_{ij} (\Ktxx)_{ij}^2
    \right]
\\&= \frac{1}{(n)_3} \left[
      \one\tp \Ktxx \Ktxx \one
    - \norm{\Ktxx}_F^2
    \right]
\\&= \frac{1}{(n)_3} \left[
      \norm{\Ktxx \one}^2
    - \norm{\Ktxx}_F^2
    \right]
,\end{align*}
\begin{align*}
    \inner{\muY}{C_X \muY}
  &\approx
    \frac{1}{n^2 (n-1)} \sum_i \sum_j \sum_{\ell \ne j} k(X_i, Y_j) k(X_i, Y_\ell)
\\&= \frac{1}{n^2 (n-1)} \left[
      \sum_{i j \ell} (\Kxy)_{ij} (\Kxy)_{i\ell}
    - \sum_{i j} (\Kxy)_{ij}^2
    \right]
\\&= \frac{1}{n^2 (n-1)} \left[
      \norm{\Kxy \one}^2
    - \norm{\Kxy}_F^2
    \right]
,\end{align*}
\begin{align*}
    \inner{\muX}{C_X \muY}
  &\approx
    \frac{1}{n^2 (n-1)} \sum_i \sum_{j \ne i} \sum_{\ell} k(X_i, X_j) k(X_i, Y_\ell)
\\&= \frac{1}{n^2 (n-1)} \sum_{i j \ell} (\Ktxx)_{ij} (\Kxy)_{i \ell}
\\&= \frac{1}{n^2 (n-1)} \one\tp \Ktxx \Kxy \one
,\end{align*}
and
\begin{align*}
    \inner{\muY}{C_X \muZ}
  &\approx
    \frac{1}{n^3} \sum_{ij\ell} k(X_i, Y_j) k(X_i, Z_\ell)
   = \frac{1}{n^3} \one\tp \Kxy\tp \Kxz \one
.\end{align*}

Finally, for the squared kernel terms:
\begin{align*}
    \E\left[ k(X_1, X_2)^2 \right]
  &\approx \frac{1}{n (n-1)} \sum_{i \ne j} k(X_i, X_j)^2
   = \frac{1}{n (n-1)} \norm{\Ktxx}_F^2
\\
    \E\left[ k(X, Y)^2 \right]
  &\approx \frac{1}{n^2} \norm{\Kxy}_F^2
.\end{align*}

\subsection{Final MMD variance estimator}
Recall the variance $V_m = \Var\left[ \mmdsqu(\setX, \setY) \right]$ of \eqref{eq:final-mmd-exp}:
\begin{align*}
    V_m
  &=
    \frac{4 (m-2)}{m (m-1)} \inner{\muX}{C_X \muX}
  - \frac{2 (2 m - 3)}{m (m-1)} \inner{\muX}{\muX}^2
\\&\qquad
  + \frac{4 (m-2)}{m (m-1)} \inner{\muY}{C_Y \muY}
  - \frac{2 (2 m - 3)}{m (m-1)} \inner{\muY}{\muY}^2
\\&\qquad
  + \frac{4 (m-2)}{m (m-1)} \Big( \inner{\muY}{C_X \muY} + \inner{\muX}{C_Y \muX} \Big)
  - \frac{4 (2 m - 3)}{m (m-1)} \inner{\muX}{\muY}^2
\\&\qquad
  - \frac{8}{m} \inner{\muX}{C_X \muY}
  + \frac{8}{m} \inner{\muX}{\muX} \inner{\muX}{\muY}
\\&\qquad
  - \frac{8}{m} \inner{\muY}{C_Y \muX}
  + \frac{8}{m} \inner{\muY}{\muY} \inner{\muX}{\muY}
\\&\qquad
  + \frac{2}{m (m-1)} \E\left[ k(X_1, X_2)^2 \right]
  + \frac{2}{m (m-1)} \E\left[ k(Y_1, Y_2)^2 \right]
  + \frac{4}{m (m-1)} \E\left[ k(X, Y)^2 \right]
.\end{align*}
Plugging in the estimators of \cref{sec:sub-estimators}, we at last get an estimator for the variance:
\begin{align*}
    \hat V_m
  &=
    \frac{4(mn + m -2n)}{(m)_2 (n)_4} \left[
        \norm{\Ktxx \one}^2 + \norm{\Ktyy \one}^2    
    \right]
\\&\qquad
    - \frac{2(2m-n)}{mn(m-1)(n-2)(n-3)}\left[ 
        \norm{\Ktxx}_F^2 + \norm{\Ktyy}_F^2    
    \right]
\\&\qquad
  + \frac{4(mn + m - 2n - 1)}{(m)_2n^2(n-1)^2}\left[ 
    \norm{\Kxy \one}^2 + \norm{\Kxy\tp \one}^2
  \right]
\\&\qquad
  - \frac{4(2m - n - 2)}{(m)_2n(n-1)^2}
    \norm{\Kxy}_F^2
  - \frac{2(2m-3)}{(m)_2(n)_4} \left[
      (\one\tp\Ktxx\one)^2 + (\one\tp\Ktyy\one)^2
    \right]
\\&\qquad
  - \frac{4(2m-3)}{(m)_2n^2(n-1)^2} (\one\tp\Kxy\one)^2
  - \frac{8}{mn^2(n-1)} \left[
      \one\tp\Ktxx\Kxy\one + \one\tp\Ktyy\Kxy\tp\one
    \right]
\\&\qquad
  + \frac{8}{mn(n)_3} \left[ 
    (\one\tp\Ktxx\one + \one\tp\Ktyy\one)(\one\tp\Kxy\one)
  \right]
\\&\qquad
  - \frac{16}{mn(n)_3} \left[ 
    \one\tp\Ktxx\Kxy\one + \one\tp\Ktyy\Kxy\tp\one  
  \right]
\end{align*}
which simplifies to
\begin{equation} \label{eq:final-mmd-est}
\begin{aligned}
    \hat V_m
  &=
    \frac{4(mn + m -2n)}{(m)_2 (n)_4} \left[
        \norm{\Ktxx \one}^2 + \norm{\Ktyy \one}^2    
    \right]
\\&\qquad
    - \frac{2(2m-n)}{mn(m-1)(n-2)(n-3)}\left[ 
        \norm{\Ktxx}_F^2 + \norm{\Ktyy}_F^2    
    \right]
\\&\qquad
  + \frac{4(mn + m - 2n - 1)}{(m)_2n^2(n-1)^2}\left[ 
    \norm{\Kxy \one}^2 + \norm{\Kxy\tp \one}^2
  \right]
\\&\qquad
  - \frac{4(2m - n - 2)}{(m)_2n(n-1)^2}
    \norm{\Kxy}_F^2
  - \frac{2(2m-3)}{(m)_2(n)_4} \left[
      (\one\tp\Ktxx\one)^2 + (\one\tp\Ktyy\one)^2
    \right]
\\&\qquad
  - \frac{4(2m-3)}{(m)_2n^2(n-1)^2} (\one\tp\Kxy\one)^2
  - \frac{8}{m(n)_3} \left[
      \one\tp\Ktxx\Kxy\one + \one\tp\Ktyy\Kxy\tp\one
    \right]
\\&\qquad
  + \frac{8}{mn(n)_3} \left[ 
    (\one\tp\Ktxx\one + \one\tp\Ktyy\one)(\one\tp\Kxy\one)
  \right]
.\end{aligned}
\end{equation}

\subsection{Final difference of MMD variance estimator}
The result \eqref{eq:final-diff-exp} of \cref{sec:diff-expression} was that
$\Var\left[ \mmdsqu(\setX, \setY) - \mmdsqu(\setX, \setZ) \right]$
is
\begin{align*}
     \nu_m
  &= 
    \frac{4 (m-2)}{m (m-1)} \left[
      \inner{\muY}{C_X \muY}
    + \inner{\muZ}{C_X \muZ}
    + \inner{\muX}{C_Y \muX}
    + \inner{\muX}{C_Z \muX}
    \right]
\\&\qquad
  + \frac{4 (m-2)}{m (m-1)} \left[
      \inner{\muY}{C_Y \muY}
    + \inner{\muZ}{C_Z \muZ}
    \right]
\\&\qquad
  - \frac{4 (2m-3)}{m (m-1)} \left[
      \inner{\muX}{\muY}^2
    + \inner{\muX}{\muZ}^2
    \right]
  - \frac{2 (2m-3)}{m (m-1)} \left[
      \inner{\muY}{\muY}^2
    + \inner{\muZ}{\muZ}^2
    \right]
\\&\qquad
  + \frac{8}{m} \Big(
      \inner{\muX}{\muY} \inner{\muX}{\muZ}
    + \inner{\muX}{\muY} \inner{\muY}{\muY}
    + \inner{\muX}{\muZ} \inner{\muZ}{\muZ}
  \Big)
\\&\qquad
  - \frac{8}{m} \Big(
      \inner{\muY}{C_X \muZ}
    + \inner{\muX}{C_Y \muY}
    + \inner{\muX}{C_Z \muZ}
    \Big)
\\&\qquad
  + \frac{4}{m (m-1)} \left(
      \E\left[ k(X, Y)^2 \right]
    + \E\left[ k(X, Z)^2 \right]
    \right)
  + \frac{2}{m (m-1)} \left(
      \E\left[ k(Y_1, Y_2)^2 \right]
    + \E\left[ k(Z_1, Z_2)^2 \right]
    \right)
,\end{align*}
which gives us the estimator
\begin{align*}
     \nu_m
  &= 
    \frac{4 (m-2)}{m (m-1) n^2(n-1)} \left[
      \norm{\Kxy \one}^2
    + \norm{\Kxy\tp \one}^2
    + \norm{\Kxz \one}^2
    + \norm{\Kxz\tp \one}^2
    - 2 \norm{\Kxy}_F^2
    - 2 \norm{\Kxz}_F^2
    \right]
\\&\qquad
  + \frac{4(m-2)}{m(m-1)(n)_3} \left[
      \norm{\Ktyy \one}^2
    - \norm{\Ktyy}_F^2
    + \norm{\Ktzz \one}^2
    - \norm{\Ktzz}_F^2
    \right]
\\&\qquad
  - \frac{4 (2m-3)}{m(m-1)n^2(n-1)^2} \left[
      \left(\one\tp \Kxy \one\right)^2
    + \left(\one\tp \Kxz \one\right)^2
    - \norm{\Kxy \one}^2
    - \norm{\Kxz \one}^2
    - \norm{\Kxy\tp \one}^2
    - \norm{\Kxz\tp \one}^2
\right.\\&\qquad\qquad\qquad\qquad\qquad\left.
    + \norm{\Kxy}_F^2
    + \norm{\Kxz}_F^2
    \right]
\\&\qquad
  - \frac{2 (2m-3)}{(m)_2 (n)_4} \left[
      \left( \one\tp \Ktyy \one\right)^2
    + \left( \one\tp \Ktzz \one\right)^2
    - 4 \norm{\Ktyy \one}^2
    - 4 \norm{\Ktzz \one}^2
    + 2 \norm{\Ktyy}_F^2
    + 2 \norm{\Ktzz}_F^2
    \right]
\\&\qquad
  + \frac{8}{m n^3(n-1)} \left[
      \one\tp \Kxy \one \one\tp \Kxz \one
    - \one\tp \Kxy\tp \Kxz \one
    \right]
\\&\qquad
  + \frac{8}{m n(n)_3} \left[
      \one\tp \Ktyy \one \one\tp \Kxy \one
    + \one\tp \Ktzz \one \one\tp \Kxz \one
    - 2 \, \one\tp \Ktyy \Kxy\tp \one
    - 2 \, \one\tp \Ktzz \Kxz\tp \one
  \right]
\\&\qquad
  - \frac{8}{mn^3} \one\tp \Kxy\tp \Kxz \one
\\&\qquad
  - \frac{8}{m n^2(n-1)} \left[
      \one\tp \Ktyy \Kxy\tp \one
    + \one\tp \Ktzz \Kxz\tp \one
    \right]
\\&\qquad
  + \frac{4}{m(m-1)n^2} \left(
      \norm{\Kxy}_F^2
    + \norm{\Kxz}_F^2
    \right)
  + \frac{2}{m(m-1)n(n-1)} \left(
      \norm{\Ktyy}_F^2
    + \norm{\Ktzz}_F^2
    \right)
.\end{align*}
Combining like terms, we obtain
\begin{equation} \label{eq:final-diff-est}
\begin{aligned}
     \hat\nu_m
  &=
    \frac{4 (mn + m -2n - 1)}{m(m-1)n^2(n-1)^2} \left[
      \norm{\Kxy \one}^2
    + \norm{\Kxy\tp \one}^2
    + \norm{\Kxz \one}^2
    + \norm{\Kxz\tp \one}^2
    \right]
\\&\qquad
  + \frac{4(mn+m-2n)}{(m)_2(n)_4} \left[
      \norm{\Ktyy \one}^2
    + \norm{\Ktzz \one}^2
    \right]
  - \frac{8}{mn^2 (n-1)} \one\tp \Kxy\tp \Kxz \one
\\&\qquad
  - \frac{8}{m(n)_3} \left[
      \one\tp \Ktyy \Kxy\tp \one
    + \one\tp \Ktzz \Kxz\tp \one
    \right]
\\&\qquad
  - \frac{4 (2m-3)}{(m)_2n^2(n-1)^2} \left[
      \left(\one\tp \Kxy \one\right)^2
    + \left(\one\tp \Kxz \one\right)^2
    \right]
  - \frac{2 (2m-3)}{(m)_2 (n)_4} \left[
      \left( \one\tp \Ktyy \one\right)^2
    + \left( \one\tp \Ktzz \one\right)^2
    \right]
\\&\qquad
  + \frac{8}{mn^3(n-1)} \one\tp \Kxy \one \one\tp \Kxz \one
  + \frac{8}{mn (n)_3} \left[
      \one\tp \Ktyy \one \one\tp \Kxy \one
    + \one\tp \Ktzz \one \one\tp \Kxz \one
    \right]
\\&\qquad
  - \frac{4 (2m-n-2)}{(m)_2 n(n-1)^2} \left[
      \norm{\Kxy}_F^2
    + \norm{\Kxz}_F^2
    \right]
  - \frac{2(2m - n)}{(m)_2 n(n-2)(n-3)} \left[
      \norm{\Ktyy}_F^2
    + \norm{\Ktzz}_F^2 
    \right]
.\end{aligned}
\end{equation}

\printbibliography

\end{document}